\documentclass{article}
\usepackage{iclr2027_conference,times}

\usepackage{amsmath,amsfonts,bm}
\usepackage{svg}

\def\eqref#1{equation~\ref{#1}}

\def\1{\bm{1}}

\DeclareMathAlphabet{\mathsfit}{\encodingdefault}{\sfdefault}{m}{sl}
\SetMathAlphabet{\mathsfit}{bold}{\encodingdefault}{\sfdefault}{bx}{n}

\usepackage{booktabs,multirow,tabularx,array}
\usepackage{hyperref}
\usepackage{url}
\usepackage{booktabs}
\usepackage{multirow}
\usepackage{graphicx}
\usepackage{xspace}
\usepackage{amsmath,amssymb}
\usepackage[ruled,vlined,linesnumbered]{algorithm2e}
\usepackage{xcolor}
\usepackage[T1]{fontenc}
\usepackage{listings}
\usepackage{booktabs,multirow,graphicx,wrapfig,caption}
\usepackage{xcolor}
\usepackage{wrapfig}
\usepackage{listings}
\usepackage[most]{tcolorbox}
\definecolor{myblue}{RGB}{0, 0, 200}
\definecolor{bgblue}{RGB}{245, 247, 255}
\definecolor{revisionpurple}{RGB}{0, 0, 0}

\newcommand{\ModelName}{\textrm{MaC}\xspace}

\title{Memory as Code: Declarative Retrieval over \\ Codified Long-Term Memory}

\author{
\textbf{Aoting Zeng\textsuperscript{1}, Yuxin Du\textsuperscript{1}, Jianwei Wang\textsuperscript{2}, Yizhang He\textsuperscript{1}} \\
\textbf{Shan Jiang\textsuperscript{1}, Xuemin Lin\textsuperscript{3}, Kai Wang\textsuperscript{1}} \\[3pt]
\small \textsuperscript{1}Shanghai Jiao Tong University \\
\small \textsuperscript{2}University of New South Wales \\
\small \textsuperscript{3}The Chinese University of Hong Kong, Shenzhen \\[3pt]
\fontsize{8}{10}\selectfont\texttt{\{aotingzeng,dyx20031122,yizhang.he,greatjs,w.kai\}@sjtu.edu.cn} \\
\fontsize{8}{10}\selectfont\texttt{jianwei.wang1@unsw.edu.au, xueminlin@cuhk.edu.cn}
}

\iclrfinalcopy
\begin{document}

\maketitle
\lhead{}

\begin{abstract}
In long-term memory systems, memories are constructed from interaction histories and retrieved in response to queries, enhancing the reasoning ability of LLM agents.
In complex scenarios, facts are scattered across sessions and their relationships evolve over time, making it difficult for existing memory systems to compose them to adapt to diverse queries. 
Motivated by the modularity and composability of code, we propose \textbf{\underline{M}}emory \textbf{\underline{a}}s \textbf{\underline{C}}ode (\ModelName), a unified memory framework that jointly designs memory construction and retrieval through a shared code abstraction. \ModelName couples \emph{codified construction}, which establishes executable memory abstractions, with \emph{declarative retrieval}, which translates queries into structured retrieval programs.
Specifically, \emph{codified construction} builds a three-layer executable memory structure that progresses from an atomic function layer, through a relational layer, to a modular composition layer. 
\emph{Declarative retrieval} recasts retrieval from relevance-based fact selection into declarative query processing over memory. 
Experiments on \textsc{LoCoMo} and \textsc{LongMemEval} show relative gains of up to 15.91\% in overall LLM-judge accuracy over the strongest baseline in each setting, while substantially reducing token consumption, highlighting the effectiveness of memory as code for long-horizon memory tasks.

\end{abstract}

\section{Introduction}
\label{sec:introduction}

Long-term memory enables Large Language Model (LLM) agents to retain and use information across interactions, supporting personalized assistance and long-horizon tasks~\citep{memorybank,generative}.
A memory system~\cite{structmem,memgpt} comprises memory construction, which organizes interaction histories into memory data, and memory retrieval, which processes memory queries over memory data. 
For memory construction, MemoryOS~\citep{memoryos} organizes dialogue into topic segments and summaries, Mem0~\citep{mem0} consolidates facts through update operations, and A-Mem~\citep{a-mem} constructs dynamically linked memory notes. These methods support memory organization and maintenance, but provide limited support for encapsulating computations over stored facts for reuse across queries. 
For memory retrieval, existing approaches combine search with contextual or relational expansion to gather relevant evidence~\citep{borgeaud2022improving,liu2023think}, using either predefined pipelines~\citep{graphrag,zep} or agentic workflows~\citep{react,searcho1,mragent}. However, predefined pipelines offer limited flexibility for diverse queries, whereas stepwise agentic methods achieve this flexibility through repeated LLM planning, incurring additional inference overhead\citep{kim2023llm,tan2025prospect}.

The above observations motivate a joint design in which memory construction encapsulates reusable computations over stored facts, and retrieval composes them for each query while reducing the need for repeated planning. Code provides a suitable foundation: shared references preserve consistent fact identities, functions expose reusable computations through explicit interfaces, and programs specify their composition and conditional branches before execution begins. This leads to a code-centric formulation in which memory is a library of callable computations and retrieval is a query-specific program over that library.

However, achieving these goals through a shared code abstraction poses two challenges:
(1) How can interaction histories be transformed into executable memory code? Facts expressed across sessions may overlap, supplement one another, or supersede earlier states, distinctions that are not automatically resolved by encoding facts as functions. Preserving consistent fact identities and distinguishing current from historical states is a key challenge.
(2) How can retrieval compose memory code to accommodate diverse queries? Relevant functions alone do not specify how their outputs should be combined, and subsequent operations may depend on intermediate results. Capturing these dependencies and conditional continuations in a complete retrieval program while reducing repeated LLM planning is another key challenge.

\begin{figure}[t]
    \centering
    \includegraphics[width=\textwidth]{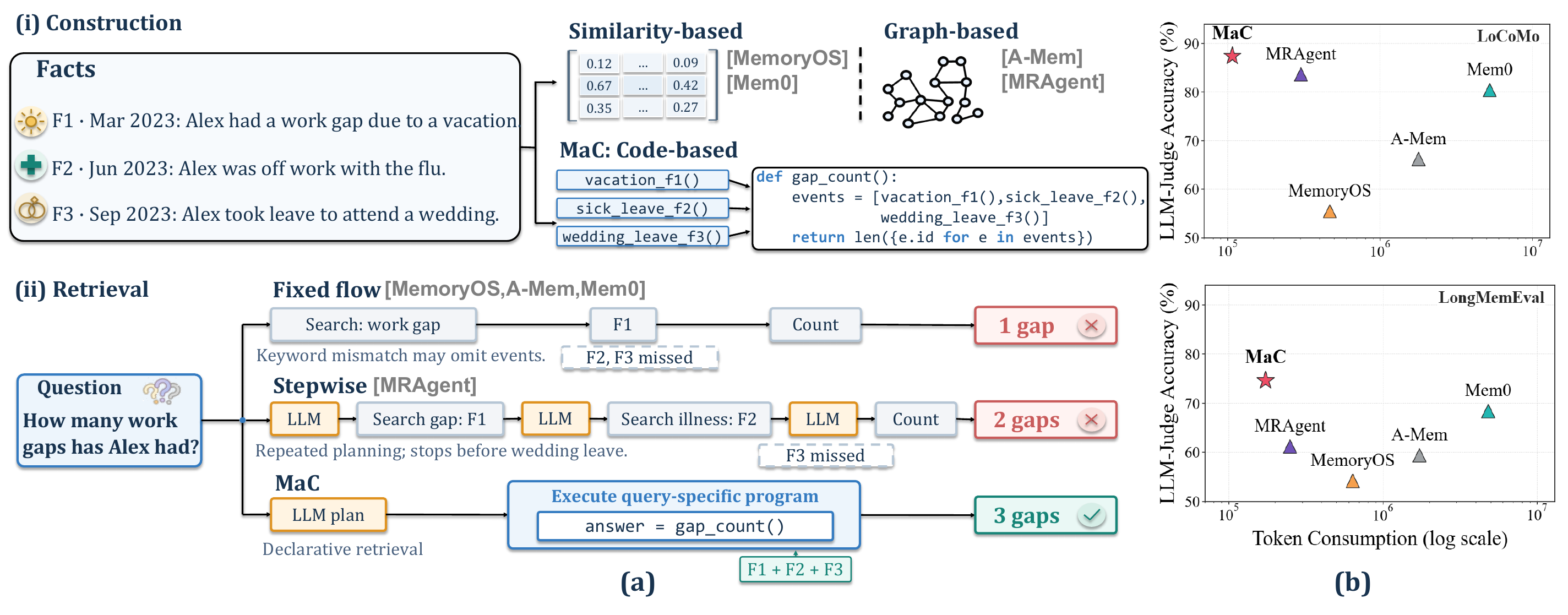}
    \caption{(a) An example illustrating the limitations of relevant methods.
    (b) MaC achieves the highest LLM-judge accuracy with the lowest token consumption among the compared methods on both \textsc{LoCoMo} and \textsc{LongMemEval}.
    }
    \label{fig:intro-limitation}
    \vspace{-2em}
\end{figure}

To address these challenges, we propose \textbf{\underline{M}}emory \textbf{\underline{a}}s \textbf{\underline{C}}ode (\ModelName), a framework that represents long-term memory as executable code and retrieves information through query-specific programs. The framework comprises two components:
(1) \emph{codified construction} transforms facts extracted from interaction histories into a three-level executable memory structure. The atomic function layer encodes facts as functions with canonical identifiers and source references; the relational layer connects fact functions through explicit supersession relations while retaining historical evidence; and the modular composition layer composes base functions into reusable current-state and history computations. A compact memory index exposes function capabilities, arguments, and available operators for retrieval.  
(2) \emph{Declarative retrieval} interprets each query as a high-level specification and compiles it into a complete retrieval program in one LLM planning call. Using the memory index, the program selects and composes the required memory computations, specifying operator arguments, dependencies, and conditional continuations. An executor checks and runs the program, following predefined branches based on intermediate results and assembling supporting evidence without stepwise LLM planning during successful execution.
The two components work together through the memory index and a common evidence representation. The index connects constructed capabilities to query requirements, while the evidence representation supports result composition and source tracing. As illustrated in Figure~\ref{fig:intro-limitation}, this design connects facts distributed across sessions through reusable functions to answer the query.

Our main contributions are as follows:
\begin{itemize}

\item We study long-term memory for LLM agents to support reasoning over evolving information across interactions. We propose  \textbf{\underline{M}}emory \textbf{\underline{a}}s \textbf{\underline{C}}ode (\ModelName), a framework that uses a shared code abstraction to connect memory capabilities with query-specific composition.

\item We develop two complementary techniques: \emph{codified construction} organizes facts, their evolving relations, and reusable computations in a three-level executable memory structure with consistent state semantics and source provenance; \emph{declarative retrieval} compiles each query into a complete retrieval program that selects and composes these capabilities in one LLM planning call.

\item Extensive experiments on widely used memory benchmarks show that \ModelName achieves the highest overall LLM-judge accuracy in all four evaluation settings, improving over the strongest baseline by up to 15.91\% on \textsc{LoCoMo} and 9.06\% on \textsc{LongMemEval}. Ablation studies further support the benefits of both techniques.
\end{itemize}

\begin{figure}[htbp]
    \centering
    \includegraphics[width=\linewidth]{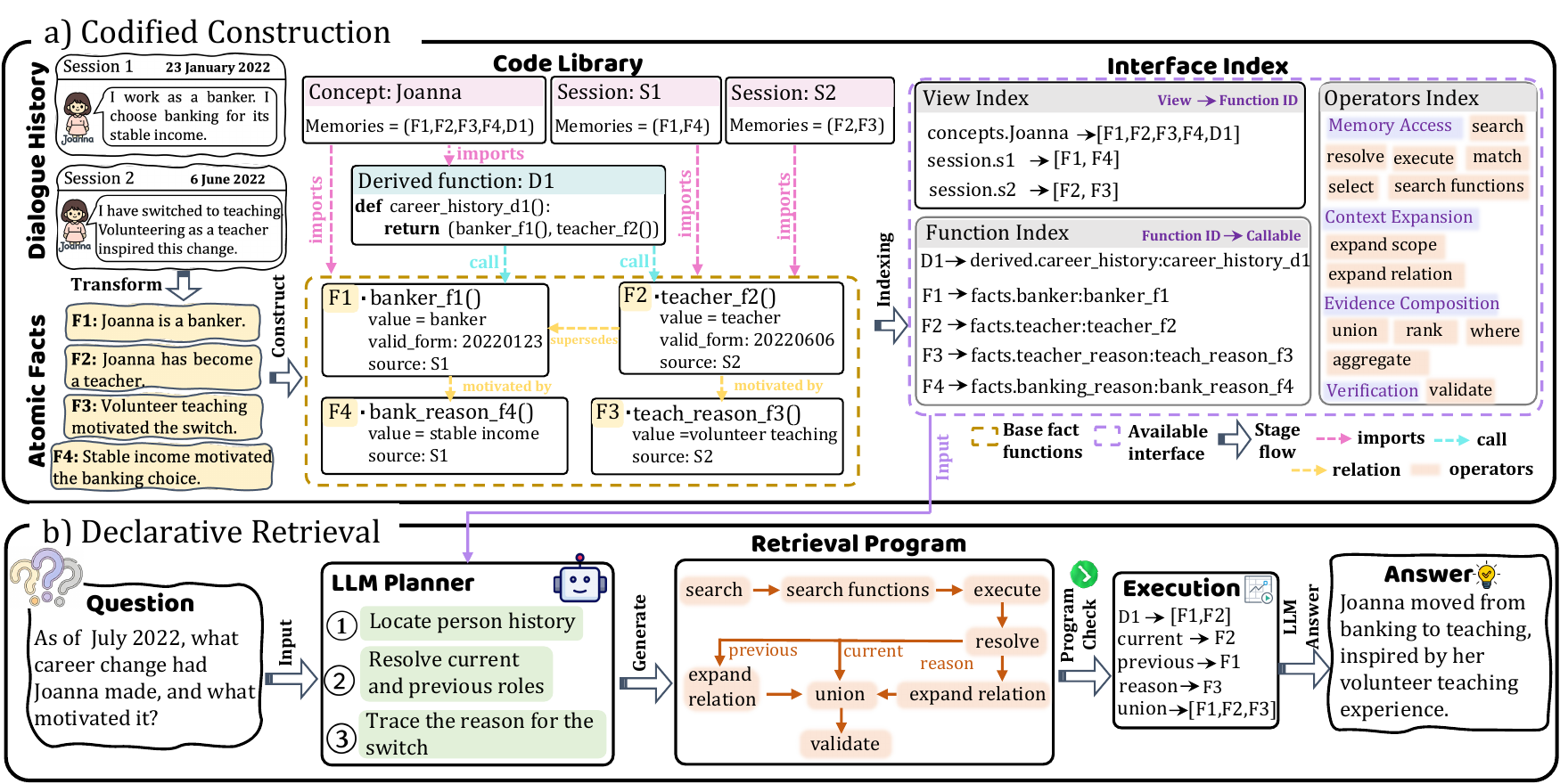}
    \caption{Overview of \ModelName. (a)~\emph{Codified construction} extracts atomic facts from dialogue history and builds a \emph{code library} of fact functions, relations, derived functions, and concept and session views, along with an \emph{interface index} that maps views to function identifiers (view index), identifiers to callable entry points (function index), and lists available memory operations (operators index). (b)~\emph{Declarative retrieval} passes the question and indexed interfaces to an \emph{LLM planner}, which generates a \emph{retrieval program}; after checking, an executor runs the program over the library to assemble supporting evidence for the answer LLM.}
    \label{fig:framework}
\end{figure}

\section{Memory as Code}
\label{sec:method}
In this section, we present \ModelName for long-term agent memory.
We first introduce the problem definition and solution framework (\S\ref{sec:overview}), then describe codified construction (\S\ref{sec:memory-construction}) and declarative retrieval (\S\ref{sec:memory-retrieval}), and finally analyze the time and space complexity (\S\ref{sec:complexity}).

\subsection{Problem Definition and Solution Framework}
\label{sec:overview}

\noindent\textbf{Problem Definition.}
Given a dialogue history $\mathcal{H}=(s_1,\ldots,s_{n_{\mathrm{sess}}})$ of $n_{\mathrm{sess}}$ sessions, memory construction organizes its information into an external memory $\mathcal{M}$.
For a query $q$, memory retrieval accesses and combines information in $\mathcal{M}$ to produce supporting evidence $E_q$, which is used to generate an answer $a_q$.
Appendix~\ref{app:notation} summarizes the notation used throughout the paper.

\noindent\textbf{Solution Framework Overview.}
\ModelName organizes memory construction and retrieval around a shared
code abstraction: memory provides reusable functions, and retrieval
composes these functions into query-specific computations.
As illustrated in Figure~\ref{fig:framework}, the framework comprises
two stages.
\emph{Codified construction} transforms dialogue history $\mathcal{H}$
into a code library of fact functions and reusable computations over
current and historical states.
A memory index exposes the callable interfaces and the available
memory operators.
\emph{Declarative retrieval} uses these interfaces to translate a
query $q$ into a complete retrieval program in one LLM planning call.
After checking, the program executes over the library to assemble
supporting evidence $E_q$ for answer generation.
The two stages are connected by indexed interfaces and a common
evidence representation, allowing retrieval to discover and compose
the capabilities established during construction.

\subsection{Codified Construction: Fact-to-Function}
\label{sec:memory-construction}
Given dialogue history $\mathcal{H}$, memory construction begins by extracting atomic facts to distill salient information and filter out irrelevant conversational details.
We then encode these facts as executable memory to support reusable computation.
However, directly encoding these facts into code is challenging because they are expressed in natural language, and the relations among them are complex and evolve over time.
To address these challenges, we design codified construction to organize facts into callable functions, explicitly represent their relations, and compose them into reusable operators.
We first introduce the codified memory representation (\S\ref{sec:compositional-memory}), then describe memory organization and indexing (\S\ref{sec:construction-procedure}).
\subsubsection{Codified Memory Representation}
\label{sec:compositional-memory}

We define \emph{codified memory} $\mathcal{M}$ that includes a \emph{code library} $\mathcal{L}$ and a \emph{memory index} $I$.
The \emph{code library} comprises four sets: \emph{base fact functions} $\mathcal{F}$, \emph{memory views} $\mathcal{V}$, \emph{fact relations} $\mathcal{R}$, and \emph{derived functions} $\mathcal{D}$:
\[
    \mathcal{M}=(\mathcal{L},I),
    \qquad
    \mathcal{L}=(\mathcal{F},\mathcal{V},\mathcal{R},\mathcal{D}).
\]

The library organizes memory into three layers.
The \emph{atomic function layer} represents individual facts as callable functions in $\mathcal{F}$.
The \emph{relational layer} specifies fact connections and state replacements through $\mathcal{R}$.
The \emph{modular composition layer} provides derived functions in $\mathcal{D}$ for reusable computations over facts and their relations, including current-state resolution and history retrieval.
Base and derived functions return evidence with fact identifiers and source references, supporting composition and provenance.

The set of memory views $\mathcal{V}$ contains \emph{concept views}, which group function references by person or topic, and \emph{session views}, which group them by dialogue session.
The memory index $I$ specifies view membership, callable function interfaces, and available memory operators, enabling retrieval to select and compose memory computations.

\subsubsection{Memory Organization and Indexing}
\label{sec:construction-procedure}

We construct the three layers of the code library $L$ from extracted facts, then build the memory index $I$ to expose the resulting functions and memory operators for retrieval.

\paragraph{Code Library Organization.}
\label{sec:code-library}
To translate natural-language facts into executable memory, we use an LLM to produce structured function specifications. A structured prompt specifies the required fact fields and function interface, and a local generator converts each specification into an executable base fact function.

At the \emph{atomic function layer}, we canonicalize and deduplicate facts and assign each a canonical identifier $i$. For each canonical fact, the LLM produces a specification containing its structured value $v_i$, including the subject, predicate, and available time information, together with a statefulness annotation. The local generator uses this specification and the supporting dialogue references $S_i$ to create a base fact function $f_i \in F$ that returns an evidence record $e_i = (v_i, \{i\}, S_i)$. Concept views and session views import shared references to these functions.

At the \emph{relational layer}, we group stateful facts by subject and predicate and ask the LLM to identify explicit replacements.
Each replacement becomes a supersession relation $(j,\texttt{supersedes},i)\in\mathcal{R}$, indicating that fact $j$ replaces fact $i$ while both functions remain available.
We also connect facts by temporal order, shared subject, and shared topic.

At the \emph{modular composition layer}, we instantiate derived functions in $\mathcal{D}$ from reusable current-state and history templates over linked facts and register them in the relevant views.
The resulting functions invoke base fact functions in $\mathcal{F}$ and use their relations in $\mathcal{R}$ to compute results.
For a state group $G$ sharing a subject and predicate, let $G(\tau)$ contain the identifiers of facts whose recorded times do not exceed the time boundary $\tau$.
The current-state template computes the active facts $A_G(\tau)$ and resolves the current value:
\begin{equation}
\begin{aligned}
    A_G(\tau)
    &=\left\{i\in G(\tau)\;\middle|\;
        \not\exists j\in G(\tau): (j,\texttt{supersedes},i)\in\mathcal{R}\right\},\\
    \operatorname{current}_G(\tau)
    &=\begin{cases}
        v_i, & A_G(\tau)=\{i\},\\
        \bot, & |A_G(\tau)|\neq 1.
    \end{cases}
\end{aligned}
\label{eq:state-resolution}
\end{equation}
Here $\bot$ denotes an unresolved state when there is no unique active fact.
The history template retains all eligible facts, including superseded ones; in Figure~\ref{fig:framework}, \texttt{D1} combines the banking fact \texttt{F1} and teaching fact \texttt{F2} into Joanna's career history.
Both templates preserve supporting fact identifiers, dialogue references, and applied computation rules as provenance.

\paragraph{Memory Index Construction.}
\label{sec:memory-index}
The \emph{code library} provides executable memory, but downstream retrieval also needs to discover its contents and determine how to use them.
We therefore construct the \emph{memory index} $I$ with two parts: indices over library contents and interfaces to memory operators.

For the library contents, we build indices at two levels.
The \emph{function index} $I_{\mathrm{func}}$ maps each base or derived function identifier to its entry point, description, and accepted arguments, extracted from function metadata.
The \emph{view index} $I_{\mathrm{view}}$ maps each concept or session view in $\mathcal{V}$ to the identifiers of the functions it references.
Together, these indices support navigation from views to callable functions without inspecting their implementations.

We define four categories of memory operators for downstream computation.
The \emph{operator index} $I_{\mathrm{ops}}$ records their interfaces.
\emph{Memory access} retrieves facts and invokes functions; \emph{context expansion} follows views and relations to gather additional facts and source context; \emph{evidence composition} combines, filters, ranks, and aggregates evidence; and \emph{verification} provides heuristic checks of evidence completeness and unresolved states.
The content indices and operator interfaces together form $I=(I_{\mathrm{view}},I_{\mathrm{func}},I_{\mathrm{ops}})$, exposing both available memory resources and the operations for using them.
Table~\ref{tab:mac2-operator-classification} in
Appendix~\ref{app:memory-operators} provides detailed operator descriptions.

\subsection{Declarative Retrieval: Query-to-Program}
\label{sec:memory-retrieval}

To adapt retrieval to diverse queries, \emph{declarative retrieval} interprets each query as a high-level specification and compiles it into a structured retrieval program that selects and composes the required memory computations.
We first describe declarative query planning (\S\ref{sec:program-synthesis}), then present retrieval program execution (\S\ref{sec:conditional-disclosure}).

\noindent\textbf{Retrieval Program.}
Given a query $q$ and codified memory $\mathcal{M}=(\mathcal{L},I)$, we define a \emph{retrieval program} $P_q$ as an executable query plan over the memory operators exposed by $I$.
The program specifies operator arguments and dependencies, together with any conditional continuations.
These operators encapsulate memory access and computation, allowing the program to compose existing memory functions and process their results through predefined operations.
Executing $P_q$ over $\mathcal{M}$ produces the supporting evidence $E_q$ specified in the problem definition.

\subsubsection{Declarative Query Planning}
\label{sec:program-synthesis}
Diverse queries require retrieval plans tailored to their information needs.
A question about a current attribute may need a single function call.
Questions combining facts across sessions may require context expansion and evidence composition.
We use an LLM planner to interpret each query and reason about the required memory operations.

Initial retrieval selects a set of candidate function identifiers $K_q$ using the memory index $I$ built during construction.
Their entries in $I_{\mathrm{func}}$ and the operator interfaces in $I_{\mathrm{ops}}$ form the planning context $I_q(\mathcal{M})$.
Function entries describe what each callable computes and which arguments it accepts.
Operator interfaces specify the available operations and their input and output constraints.
This compact context supports planning without loading the full \emph{code library}.

Our structured prompt presents the query, function information, operator constraints, and planning guidance (Appendix~\ref{app:prompt-details}).
The guidance illustrates how query requirements translate into operations, such as temporal filtering or expansion across sessions.
The prompt also specifies evidence compatibility constraints and requires the program to return supporting evidence.
The LLM uses the query's constraints to select relevant functions and operators, then links their outputs to subsequent inputs.

The planner combines these inputs to produce a complete retrieval program $P_q=F_{\mathrm{plan}}(q,I_q(\mathcal{M}))$ in one planning call.
In restricted Python, it specifies operator calls, arguments, dependencies, and conditional retrieval actions based on intermediate evidence.

\subsubsection{Retrieval Program Execution}
\label{sec:conditional-disclosure}

Before execution, \ModelName checks the retrieval program's structural compatibility with the available memory operations, including permitted language constructs, operator calls, variable references, and evidence dependencies.
These dependencies connect function invocation, any selected view expansion, and subsequent evidence processing, preserving the information required by each operation.
The query determines which computations and conditional continuations the program combines.

Once accepted, $P_q$ executes over the code library $\mathcal{L}$, with function calls resolved through the memory index $I$.
Calls to base and derived functions reuse the fact definitions and state computations established during construction, while evidence composition preserves their fact identities and sources.
Intermediate results determine which conditional continuations already specified in $P_q$ are followed.
Execution can therefore perform additional memory access when indicated by these conditions without a new LLM planning decision.
This planning--execution separation combines operator composition tailored to each query with reusable memory computation.

The final evidence assembly combines selected facts, supporting dialogue, and computation provenance into $E_q$, loading source context as specified by the program; execution also produces a trace $\mathcal{T}_q$.
A separate answer model generates the response as $a_q=F_{\mathrm{answer}}(q,E_q)$.
A valid first attempt uses one planning call and a separate answer-generation call; checking or execution failures may trigger bounded repair attempts, followed by a fixed retrieval fallback if needed.
Algorithm~\ref{alg:mac-framework} in Appendix~\ref{app:framework-algorithm} summarizes construction, retrieval, and answer generation using the same notation.

\subsection{Complexity Analysis}
\label{sec:complexity}
One-time construction costs $T_{\mathrm{build}}=O(h\log h+n^2\log n+\mathcal{C}_{\mathrm{build}})$, where $n$ and $h$ are the number of extracted facts and dialogue length in tokens and $\mathcal{C}_{\mathrm{build}}$ aggregates construction-stage model inference.
Each query costs $T_q=O\bigl(\mathcal{C}_q+p(n^2+h+b^2)+p^2\bigr)$ time for a retrieval program with $p$ syntax-tree nodes and evidence expansions of size up to $b$, with $\mathcal{C}_q$ capturing planning and answering.
Persistent codified memory occupies $O(n+h)$ space because stored references satisfy $r=O(n)$, and query processing adds $O(pb)$ for program intermediates and execution traces.
Appendix~\ref{app:complexity} states the assumptions and gives the full time and space analysis.

\begin{table*}[t]
    \centering
    \caption{Performance across different question types on \textsc{LoCoMo}. 
    F1 is reported in percent, and LLM-Judge accuracy (J) is reported as mean ~$\pm$~standard deviation in percent over five independent judge evaluations. Improv.\ denotes the relative improvement of MaC over each corresponding baseline in Overall J. \textbf{Bold} and \underline{underline} mark the best and second-best results.}
    \label{tab:locomo-results}
    \small
    \setlength{\tabcolsep}{3pt}
    \resizebox{\textwidth}{!}{%
    \begin{tabular}{llrrrrrrrrrr@{\hspace{2pt}}r}
    \toprule
    \multirow{2}{*}{Model} & \multirow{2}{*}{Method} & \multicolumn{2}{c}{Multi-hop} & \multicolumn{2}{c}{Temporal} & \multicolumn{2}{c}{Open-domain} & \multicolumn{2}{c}{Single-hop} & \multicolumn{2}{c}{Overall} & \multirow{2}{*}{Improv.} \\
    \cmidrule(lr){3-4} \cmidrule(lr){5-6} \cmidrule(lr){7-8} \cmidrule(lr){9-10} \cmidrule(lr){11-12}
    & & \multicolumn{1}{c}{F1 $\uparrow$} & \multicolumn{1}{c}{J $\uparrow$} & \multicolumn{1}{c}{F1 $\uparrow$} & \multicolumn{1}{c}{J $\uparrow$} & \multicolumn{1}{c}{F1 $\uparrow$} & \multicolumn{1}{c}{J $\uparrow$} & \multicolumn{1}{c}{F1 $\uparrow$} & \multicolumn{1}{c}{J $\uparrow$} & \multicolumn{1}{c}{F1 $\uparrow$} & \multicolumn{1}{c}{J $\uparrow$} & \\
    \midrule
    \multirow{5}{*}{\shortstack{Qwen\\3.6-27B}}
    & A-Mem & 39.44 & 57.73\textsubscript{$\pm 2.92$} & 37.16 & 45.23\textsubscript{$\pm 3.39$} & 20.13 & 45.00\textsubscript{$\pm 0.87$} & 54.40 & 68.80\textsubscript{$\pm 1.72$} & 45.93 & 60.38\textsubscript{$\pm 0.80$} & +43.82\% \\
    & MemoryOS & 29.42 & 42.06\textsubscript{$\pm 2.17$} & 35.96 & 22.12\textsubscript{$\pm 0.58$} & 23.43 & 43.96\textsubscript{$\pm 1.71$} & 38.93 & 50.20\textsubscript{$\pm 0.96$} & 35.60 & 42.47\textsubscript{$\pm 1.03$} & +104.47\% \\
    & Mem0 & 43.67 & 71.06\textsubscript{$\pm 2.19$} & \underline{60.27} & \underline{81.68}\textsubscript{$\pm 0.14$} & \underline{30.53} & 56.88\textsubscript{$\pm 0.93$} & 53.93 & \underline{75.70}\textsubscript{$\pm 0.35$} & 51.91 & \underline{74.92}\textsubscript{$\pm 0.28$} & +15.91\% \\
    & MRAgent & \underline{48.88} & \underline{75.04}\textsubscript{$\pm 2.49$} & 58.36 & 71.03\textsubscript{$\pm 0.38$} & 28.19 & \underline{57.92}\textsubscript{$\pm 1.58$} & \underline{55.01} & 70.11\textsubscript{$\pm 0.69$} & \underline{52.92} & 70.44\textsubscript{$\pm 0.81$} & +23.28\% \\
    & \textbf{MaC} & \textbf{56.20} & \textbf{87.09}\textsubscript{$\pm 0.74$} & \textbf{65.78} & \textbf{83.93}\textsubscript{$\pm 0.87$} & \textbf{37.36} & \textbf{63.75}\textsubscript{$\pm 0.47$} & \textbf{73.49} & \textbf{90.51}\textsubscript{$\pm 0.41$} & \textbf{66.46} & \textbf{86.84}\textsubscript{$\pm 0.35$} & \multicolumn{1}{c}{--} \\
    \midrule
    \multirow{5}{*}{\shortstack{ChatGPT\\5.6-Luna}}
    & A-Mem & 41.59 & 60.43\textsubscript{$\pm 1.24$} & 54.26 & 62.55\textsubscript{$\pm 0.81$} & 25.32 & 51.46\textsubscript{$\pm 2.03$} & 53.23 & 70.06\textsubscript{$\pm 0.95$} & 49.58 & 65.57\textsubscript{$\pm 0.94$} & +33.69\% \\
    & MemoryOS & 32.45 & 58.65\textsubscript{$\pm 1.11$} & 42.22 & 38.69\textsubscript{$\pm 0.14$} & 22.01 & 50.00\textsubscript{$\pm 1.47$} & 41.64 & 68.80\textsubscript{$\pm 1.05$} & 38.85 & 59.49\textsubscript{$\pm 0.75$} & +47.35\% \\
    & Mem0 & 40.96 & 70.78\textsubscript{$\pm 3.09$} & 55.25 & 73.96\textsubscript{$\pm 0.28$} & 28.83 & 57.50\textsubscript{$\pm 1.36$} & 52.54 & \underline{88.09}\textsubscript{$\pm 0.26$} & 49.51 & 80.06\textsubscript{$\pm 0.69$} & +9.49\% \\
    & MRAgent & \underline{51.86} & \underline{71.56}\textsubscript{$\pm 1.51$} & \underline{67.64} & \underline{76.82}\textsubscript{$\pm 1.28$} & \underline{38.60} & \underline{60.21}\textsubscript{$\pm 1.71$} & \textbf{69.15} & 86.66\textsubscript{$\pm 0.50$} & \underline{63.76} & \underline{80.19}\textsubscript{$\pm 0.77$} & +9.32\% \\
    & \textbf{MaC} & \textbf{52.41} & \textbf{82.20}\textsubscript{$\pm 0.68$} & \textbf{69.81} & \textbf{86.29}\textsubscript{$\pm 0.44$} & \textbf{39.78} & \textbf{64.58}\textsubscript{$\pm 1.28$} & \underline{68.41} & \textbf{92.65}\textsubscript{$\pm 0.21$} & \textbf{63.99} & \textbf{87.66}\textsubscript{$\pm 0.15$} & \multicolumn{1}{c}{--} \\
    \midrule
    \multirow{5}{*}{\shortstack{DeepSeek\\V4-Flash}}
    & A-Mem & 41.63 & 63.69\textsubscript{$\pm 1.89$} & 37.58 & 42.74\textsubscript{$\pm 2.07$} & 25.15 & 52.92\textsubscript{$\pm 1.71$} & 55.88 & 77.53\textsubscript{$\pm 1.33$} & 47.54 & 66.21\textsubscript{$\pm 1.40$} & +31.99\% \\
    & MemoryOS & 27.26 & 52.06\textsubscript{$\pm 2.72$} & 32.80 & 32.09\textsubscript{$\pm 4.70$} & 22.96 & 50.83\textsubscript{$\pm 2.00$} & 40.84 & 66.06\textsubscript{$\pm 0.96$} & 35.57 & 55.47\textsubscript{$\pm 2.11$} & +57.54\% \\
    & Mem0 & 39.44 & \underline{76.45}\textsubscript{$\pm 0.89$} & 58.09 & \underline{83.49}\textsubscript{$\pm 0.00$} & \underline{34.16} & \textbf{65.21}\textsubscript{$\pm 0.93$} & 50.54 & 82.28\textsubscript{$\pm 1.07$} & 49.06 & 80.40\textsubscript{$\pm 0.79$} & +8.69\% \\
    & MRAgent & \underline{48.02} & 76.10\textsubscript{$\pm 1.47$} & \underline{69.67} & 81.74\textsubscript{$\pm 0.28$} & 27.92 & 55.42\textsubscript{$\pm 0.87$} & \underline{66.45} & \underline{90.11}\textsubscript{$\pm 0.35$} & \underline{61.35} & \underline{83.64}\textsubscript{$\pm 0.49$} & +4.48\% \\
    & \textbf{MaC} & \textbf{52.99} & \textbf{86.81}\textsubscript{$\pm 0.30$} & \textbf{70.48} & \textbf{84.74}\textsubscript{$\pm 0.00$} & \textbf{40.01} & \underline{64.58}\textsubscript{$\pm 0.00$} & \textbf{74.07} & \textbf{91.20}\textsubscript{$\pm 0.12$} & \textbf{67.34} & \textbf{87.39}\textsubscript{$\pm 0.03$} & \multicolumn{1}{c}{--} \\
    \bottomrule
    \end{tabular}%
    }
\end{table*}
\begin{table*}[t]
    \centering
    \caption{Performance across different question types on \textsc{LongMemEval} using the Qwen backbone. F1 and LLM-Judge accuracy (J) are reported in percent. Sgl.: single-session-user; Tmp.: temporal-reasoning; Pref.: single-session-preference; Upd.: knowledge-update. Bold indicates the best result, and underline indicates the second best in each column.}
    \label{tab:longmemeval-qwen-results}
    \small
    \setlength{\tabcolsep}{3pt}
    \begin{tabular}{lrrrrrrrrrrr}
    \toprule
    \multirow{2}{*}{Method} & \multicolumn{2}{c}{Sgl.} & \multicolumn{2}{c}{Tmp.} & \multicolumn{2}{c}{Pref.} & \multicolumn{2}{c}{Upd.} & \multicolumn{2}{c}{Overall} & \multirow{2}{*}{Improv.} \\
    \cmidrule(lr){2-3} \cmidrule(lr){4-5} \cmidrule(lr){6-7} \cmidrule(lr){8-9} \cmidrule(lr){10-11}
    & \multicolumn{1}{c}{F1 $\uparrow$} & \multicolumn{1}{c}{J $\uparrow$} & \multicolumn{1}{c}{F1 $\uparrow$} & \multicolumn{1}{c}{J $\uparrow$} & \multicolumn{1}{c}{F1 $\uparrow$} & \multicolumn{1}{c}{J $\uparrow$} & \multicolumn{1}{c}{F1 $\uparrow$} & \multicolumn{1}{c}{J $\uparrow$} & \multicolumn{1}{c}{F1 $\uparrow$} & \multicolumn{1}{c}{J $\uparrow$} & \\
    \midrule
    A-Mem & 71.74 & 84.29 & 32.41 & 36.84 & 7.05 & 50.00 & 61.81 & 73.08 & 47.16 & 59.40 & +25.59\% \\
    MemoryOS & 70.69 & 80.00 & 30.63 & 32.33 & 6.43 & 63.33 & 54.99 & 69.23 & 43.33 & 54.20 & +37.64\% \\
    Mem0 & \underline{82.63} & \underline{92.86} & \underline{34.62} & 54.89 & \textbf{19.15} & \underline{70.00} & \underline{62.71} & \underline{76.92} & \underline{50.95} & \underline{68.40} & +9.06\% \\
    MRAgent & 73.87 & 88.57 & 34.03 & \underline{56.39} & 6.94 & 53.33 & 43.21 & 53.85 & 43.67 & 61.20 & +21.90\% \\
    \textbf{\ModelName} & \textbf{84.28} & \textbf{94.29} & \textbf{53.28} & \textbf{68.42} & \underline{14.49} & \textbf{73.33} & \textbf{69.69} & \textbf{87.18} & \textbf{58.58} & \textbf{74.60} & \multicolumn{1}{c}{--} \\
    \bottomrule
    \end{tabular}
\end{table*}

\section{Experiments}
\label{sec:experiments}

\subsection{Experimental Setups}
\label{sec:experimental-setups}
In this section, we describe the experimental setup and present comprehensive results to systematically evaluate \ModelName and address the following research questions (RQs):
\begin{itemize}
    \item \textbf{RQ1 (Overall performance)} How does \ModelName perform against representative long-term memory systems across benchmarks, question types, and LLM backbones?

    \item \textbf{RQ2 (Cost, runtime, and storage)} What are the token consumption, wall-clock runtime, and storage footprint of \ModelName during memory construction and retrieval?

    \item \textbf{RQ3 (Ablation study)} How do executable memory and programmatic search individually and jointly contribute to \ModelName's performance?

\end{itemize}

\noindent\textbf{Datasets.}
We evaluate \ModelName on two benchmarks for long-term conversational memory: \textsc{LoCoMo}~\citep{locomo}, which evaluates question answering over long conversations, and \textsc{LongMemEval}~\citep{longmemeval}, which evaluates memory across multiple sessions with longer interaction histories per query.

\noindent\textbf{Baselines.}
We compare \ModelName with  representative baselines: A-Mem~\citep{a-mem}, MemoryOS~\citep{memoryos}, Mem0~\citep{mem0}, and MRAgent~\cite{mragent}. 

\noindent\textbf{Implementation Details.}
We conduct our experiments on NVIDIA A100 GPUs. On \textsc{LoCoMo}, we evaluate all methods using three LLM backbones: Qwen3.6-27B, ChatGPT-5.6-Luna, and DeepSeek-V4-Flash. On \textsc{LongMemEval}, we use Qwen3.6-27B as the backbone for all methods. Following prior work, we report F1 and LLM-Judge (J) scores. LLM-Judge (J) scores are obtained using DeepSeek-V4-Flash. Our code is available at:{\url{https://github.com/zengaoting/MemoryasCode}}.

\subsection{Main Results (RQ1)}
\label{sec:main-results}
To evaluate the effectiveness of \ModelName, we compare it with four representative memory systems on \textsc{LoCoMo} and \textsc{LongMemEval}, with three generation backbones evaluated on \textsc{LoCoMo}.
Tables~\ref{tab:locomo-results} and~\ref{tab:longmemeval-qwen-results} summarize the results, from which we draw two main observations.

\noindent\textbf{Overall performance.}
Tables~\ref{tab:locomo-results} and~\ref{tab:longmemeval-qwen-results} show that \ModelName achieves the highest overall J in all four evaluation settings.
On \textsc{LoCoMo}, it outperforms the strongest baseline for each backbone by 3.75--11.92 percentage points in overall J.
On \textsc{LongMemEval}, it improves overall F1 and J by 7.63 and 6.20 points, respectively.

\noindent\textbf{Temporal reasoning and knowledge updates.}
\ModelName achieves the best temporal F1 and J across all three \textsc{LoCoMo} backbones.
On \textsc{LongMemEval}, it surpasses the strongest respective baselines by 18.66 points in temporal F1, 12.03 points in temporal J, and 10.26 points in knowledge-update J, highlighting its effectiveness on queries involving temporal constraints and evolving facts.

\subsection{Cost and Efficiency Analysis (RQ2)}
\label{sec:cost-analysis}

\begin{figure*}[t]
    \centering
    \includegraphics[width=0.49\textwidth]
    {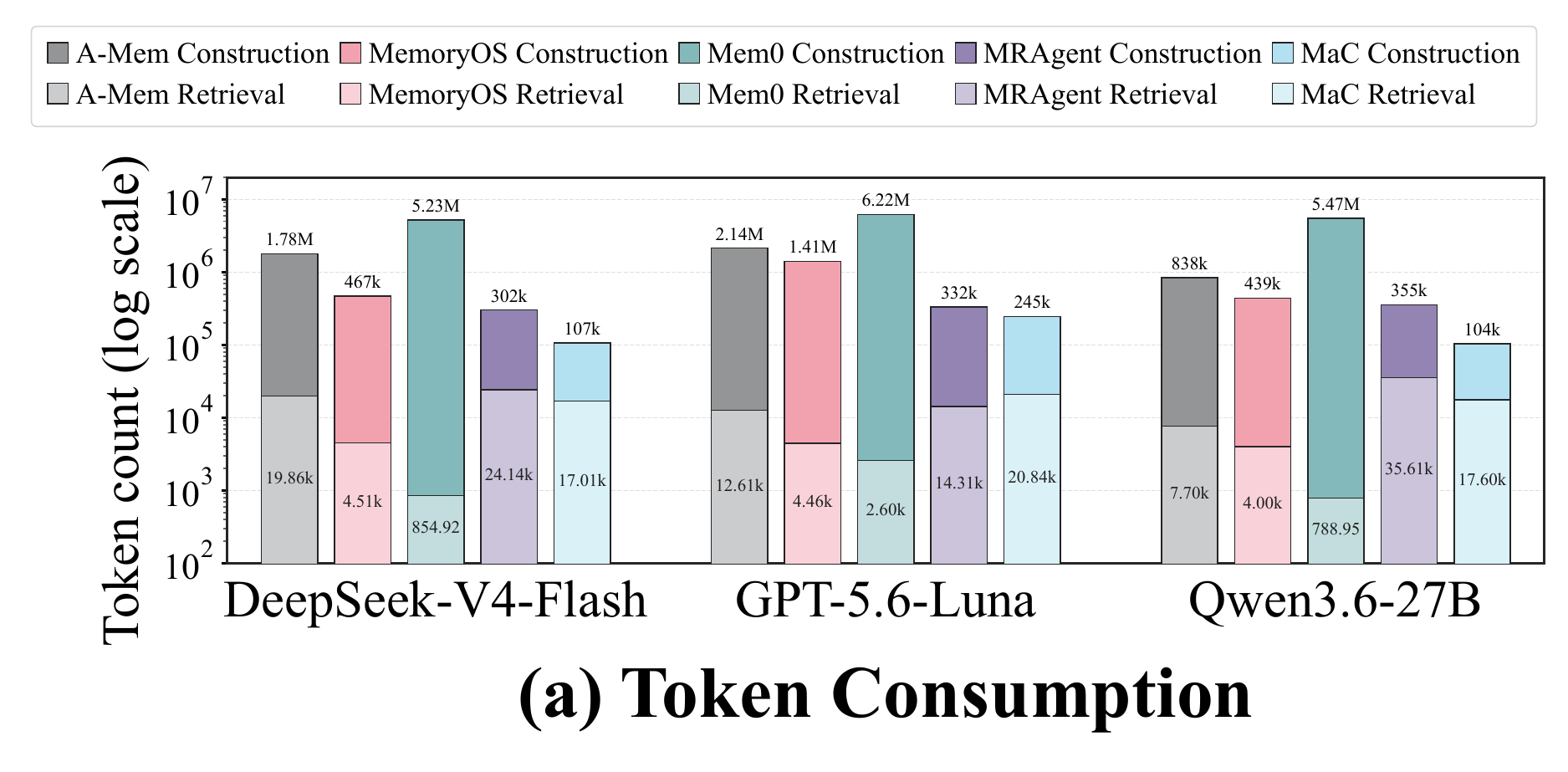}
    \hfill
    \includegraphics[width=0.49\textwidth]
    {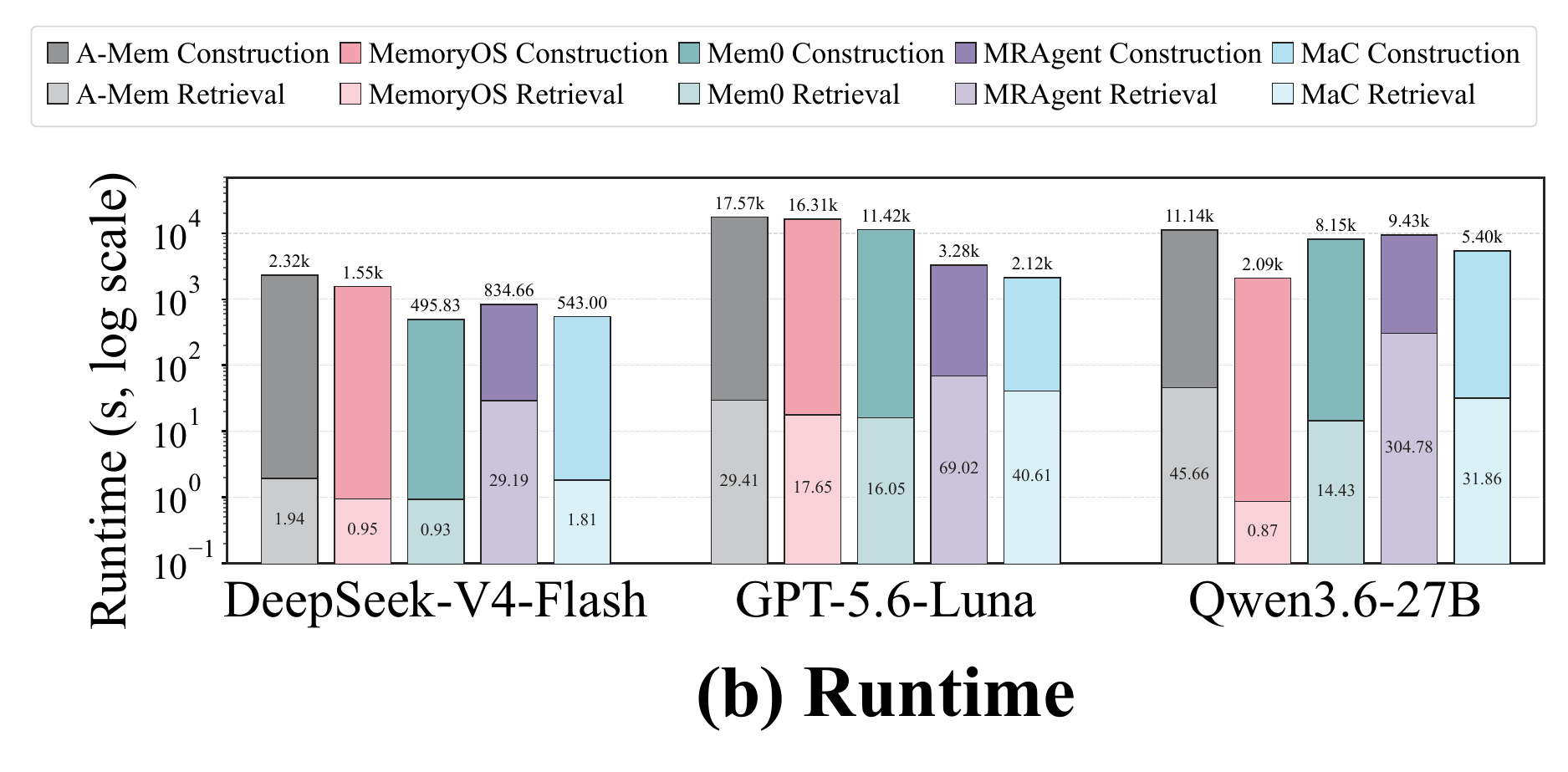}

    \caption{Normalized token consumption (a) and runtime (b)
    on \textsc{LoCoMo} across different generation models.
    Construction is normalized by 10 conversations and retrieval
    by 1,540 questions. Both panels use logarithmic vertical axes.}
    \label{fig:locomo-costs}
\end{figure*}

We report construction and query-processing costs separately on \textsc{LoCoMo} in Figure~\ref{fig:locomo-costs}, and their combined per-sample costs on \textsc{LongMemEval} in Table~\ref{tab:longmemeval-qwen-per-sample-cost}.

\noindent\textbf{Token consumption.}
Figure~\ref{fig:locomo-costs}(a) shows that \ModelName requires the fewest tokens for memory construction across all three backbones on \textsc{LoCoMo}.
This demonstrates that building an executable memory library and its interface index can be achieved with a lower construction token budget than the evaluated baselines.
The advantage is most consistent during construction, while query-stage token consumption varies across backbones and methods.
On \textsc{LongMemEval}, \ModelName also achieves the lowest combined token consumption, using approximately 173k tokens per sample compared with 249k for MRAgent, a reduction of approximately 30.5\%.

\noindent\textbf{Runtime.}
Figure~\ref{fig:locomo-costs}(b) shows that \ModelName achieves competitive runtime across all three backbones on \textsc{LoCoMo}.
It ranks first or second in memory construction time and consistently achieves lower query latency than MRAgent.
On \textsc{LongMemEval}, Table~\ref{tab:longmemeval-qwen-per-sample-cost} shows that \ModelName achieves the second-lowest end-to-end runtime among the evaluated methods, alongside the lowest token consumption.
Together with the highest overall LLM-judge accuracy reported in Table~\ref{tab:longmemeval-qwen-results}, these results show that \ModelName maintains competitive computational efficiency while providing effective memory-based reasoning.

\begin{table}[t]
    \centering
    \caption{Per-sample token consumption and runtime on \textsc{LongMemEval} across different memory systems using the Qwen backbone. Results include both memory construction and retrieval. \textbf{Bold} indicates the best result, and \underline{underline} indicates the second best.}
    \label{tab:longmemeval-qwen-per-sample-cost}
    \small
    \setlength{\tabcolsep}{5pt}
    \begin{tabular}{lrr}
    \toprule
    Method & \multicolumn{1}{c}{Token Consumption $\downarrow$} & \multicolumn{1}{c}{Runtime (s) $\downarrow$} \\
    \midrule
    A-Mem    & 1,723k & 16,417.98 \\
    MemoryOS & 636k & 11,256.63 \\
    Mem0     & 4,812k & 11,952.47 \\
    MRAgent  & \underline{249k} & \textbf{2,963.90} \\
    MAC      & \textbf{173k} & \underline{4,663.66} \\
    \bottomrule
    \end{tabular}
    \vspace{-2em}
\end{table}

\noindent\textbf{Storage space.}
As shown in Table~\ref{tab:locomo-normalized-storage-kb}, MaC maintains a competitive storage footprint among the methods evaluated on \textsc{LoCoMo}.
\ModelName requires 2,859.18\,KB of total storage, a 33.5\% reduction compared with MRAgent, a strong baseline. Although A-Mem has a smaller storage footprint (1,338.88\,KB), it achieves substantially lower accuracy than \ModelName under the same setting.

\begin{table}[t]
    \centering
    \caption{Storage space usage on \textsc{LoCoMo} (KB). Construction storage is averaged per conversation, and retrieval storage is averaged per question. \textbf{Bold} indicates the best result, and \underline{underline} indicates the second best in each column.}
    \label{tab:locomo-normalized-storage-kb}
    \small
    \setlength{\tabcolsep}{8pt}
    \begin{tabular}{cccc}
    \toprule
    Method & Construction $\downarrow$ & Retrieval $\downarrow$ & Total $\downarrow$ \\
    \midrule
    A-Mem    & \textbf{1,161.82}   & \underline{177.06} & \textbf{1,338.88} \\
    MemoryOS & 11,452.02            & 415.59             & 11,867.61 \\
    Mem0     & 7,105.06             & \textbf{13.20}     & 7,118.25 \\
    MRAgent  & 3,870.32             & 428.13             & 4,298.46 \\
    MaC      & \underline{2,565.79} & 293.39             & \underline{2,859.18} \\
    \bottomrule
    \end{tabular}
\end{table}

\subsection{Ablation Study (RQ3)}
\label{sec:ablation}
Table~\ref{tab:locomo-ablation} reports sequential ablations on \textsc{LoCoMo} with DeepSeek-V4-Flash.
MaC w/o DR retains the codified library but uses a fixed retrieval program; MaC w/o DR+CC applies that program to extracted atomic facts only.

\noindent\textbf{Effect of \emph{declarative retrieval}.}
MaC w/o DR keeps the codified library but uses a fixed retrieval program that combines lexical recall over facts with execution of candidate memory functions through predefined memory operators.
Enabling DR improves overall J by 6.62 percentage points, demonstrating the benefit of adapting retrieval operations and their composition to each query.

\noindent\textbf{Effect of \emph{codified construction}.}
Under fixed retrieval, CC improves overall J by 8.05 percentage points, supporting the value of reusable memory computations and indexed access.
Removing both DR and CC applies the same fixed program to extracted atomic facts only and reduces overall J by 14.67 percentage points relative to full \ModelName.

\begin{table*}[t]
    \centering
    \caption{Sequential ablation of declarative retrieval (DR) and codified construction (CC) on \textsc{LoCoMo} with DeepSeek-V4-Flash. F1 and binary LLM-judge accuracy (J) are percentages. Bold indicates the best result in each column. We use \textcolor{green!60!black}{$\uparrow$} to indicate an improvement in the performance of one module, and \textcolor{green!60!black}{$\uparrow\uparrow$} to indicate improvements in two modules.}
    \label{tab:locomo-ablation}
    \small
    \setlength{\tabcolsep}{3pt}
    \resizebox{\textwidth}{!}{%
    \begin{tabular}{l*{10}{l}r}
    \toprule
    \multirow{2}{*}{Method} & \multicolumn{2}{c}{Multi-hop} & \multicolumn{2}{c}{Temporal} & \multicolumn{2}{c}{Open-domain} & \multicolumn{2}{c}{Single-hop} & \multicolumn{2}{c}{Overall} & \multirow{2}{*}{Improv.} \\
    \cmidrule(lr){2-3} \cmidrule(lr){4-5} \cmidrule(lr){6-7} \cmidrule(lr){8-9} \cmidrule(lr){10-11}
    & \multicolumn{1}{c}{F1 $\uparrow$} & \multicolumn{1}{c}{J $\uparrow$} & \multicolumn{1}{c}{F1 $\uparrow$} & \multicolumn{1}{c}{J $\uparrow$} & \multicolumn{1}{c}{F1 $\uparrow$} & \multicolumn{1}{c}{J $\uparrow$} & \multicolumn{1}{c}{F1 $\uparrow$} & \multicolumn{1}{c}{J $\uparrow$} & \multicolumn{1}{c}{F1 $\uparrow$} & \multicolumn{1}{c}{J $\uparrow$} & \\
    \midrule
    MaC w/o DR+CC & 44.01 & 70.92 & 50.30 & 64.80 & 36.97 & 58.33 & 61.09 & 78.00 & 54.21 & 72.73 & +20.17\% \\
    MaC w/o DR & 50.54\,\textcolor{green!60!black}{$\uparrow$} & 85.46\,\textcolor{green!60!black}{$\uparrow$} & 59.63\,\textcolor{green!60!black}{$\uparrow$} & 69.78\,\textcolor{green!60!black}{$\uparrow$} & 39.00\,\textcolor{green!60!black}{$\uparrow$} & 62.50\,\textcolor{green!60!black}{$\uparrow$} & 67.15\,\textcolor{green!60!black}{$\uparrow$} & 85.49\,\textcolor{green!60!black}{$\uparrow$} & 60.79\,\textcolor{green!60!black}{$\uparrow$} & 80.78\,\textcolor{green!60!black}{$\uparrow$} & +8.20\% \\
    \textbf{\ModelName} & \textbf{52.99}\,\textcolor{green!60!black}{$\uparrow\uparrow$} & \textbf{86.52}\,\textcolor{green!60!black}{$\uparrow\uparrow$} & \textbf{70.48}\,\textcolor{green!60!black}{$\uparrow\uparrow$} & \textbf{84.74}\,\textcolor{green!60!black}{$\uparrow\uparrow$} & \textbf{40.01}\,\textcolor{green!60!black}{$\uparrow\uparrow$} & \textbf{64.58}\,\textcolor{green!60!black}{$\uparrow\uparrow$} & \textbf{74.07}\,\textcolor{green!60!black}{$\uparrow\uparrow$} & \textbf{91.32}\,\textcolor{green!60!black}{$\uparrow\uparrow$} & \textbf{67.34}\,\textcolor{green!60!black}{$\uparrow\uparrow$} & \textbf{87.40}\,\textcolor{green!60!black}{$\uparrow\uparrow$} & \multicolumn{1}{c}{--} \\
    \bottomrule
    \end{tabular}%
    }
\end{table*}

\section{Related Work}
\label{sec:related-work}

\noindent\textbf{Memory Construction.}
Memory construction concerns retaining useful information from growing interaction histories and revising it as new interactions arrive. A related foundation is retrieval-augmented generation (RAG)~\citep{rag}, which indexes external passages as dense vectors for generation; GraphRAG~\citep{graphrag} augments this setting with graph structure, community summaries, and neighborhood-based access. For persistent conversational memory, Mem0~\citep{mem0} extracts salient facts and reconciles them with existing memory, whereas MemoryOS~\citep{memoryos} organizes information across short-, mid-, and long-term stores. Other methods emphasize relational and temporal structure: A-Mem~\citep{a-mem} dynamically links and updates structured notes, and Zep~\citep{zep} records fact validity and historical changes in a bi-temporal graph. More recently, MemSkill~\citep{memskill} learns to select and evolve memory-construction skills. Collectively, these methods improve information selection, consolidation, and structural organization through different memory representations and construction strategies.

\noindent\textbf{Memory Retrieval.}
Memory retrieval addresses both locating relevant records and deciding how further evidence should be accessed and composed. SeCom~\citep{secom} improves retrieved context through topic-based segmentation and compression-based denoising. Structured retrieval extends access beyond flat text matching: Zep~\citep{zep} combines semantic and full-text search with graph traversal and reranking. In external-knowledge search, Search-o1~\citep{searcho1} integrates on-demand search into document-based reasoning, and Search-R1~\citep{searchr1} uses reinforcement learning to learn multi-turn search behavior. At the execution-interface level, Text2Mem~\citep{text2mem} translates memory instructions into typed, validated operations and supports workflows with dependencies across backends. MRAgent~\citep{mragent} uses intermediate evidence to guide exploration and pruning over a Cue--Tag--Content graph. These approaches improve context quality, structured access, or adaptive exploration; nevertheless, candidate-based retrieval remains bounded by the records selected for access, while iterative reasoning can require repeated context processing and model calls. Thus, retrieving sufficient evidence under a limited budget remains a central challenge.

\section{Conclusion}
\label{sec:conclusion}
We presented \ModelName, a unified framework that jointly designs memory construction and retrieval. We develop two complementary techniques:
\emph{codified construction} organizes facts, their evolving relations, and reusable computations into a three-layer executable memory structure.
\emph{declarative retrieval} compiles queries into complete programs that select and compose these computations, enabling query-specific conditional execution without stepwise LLM planning.
Experiments on \textsc{LoCoMo} and \textsc{LongMemEval} show that \ModelName achieves the highest overall F1 and LLM-judge scores across the evaluated settings while substantially reducing token consumption.

\bibliography{iclr2027_conference}
\bibliographystyle{iclr2027_conference}

\appendix
\section{Appendix}
\label{sec:appendix}

\subsection{Notation}
\label{app:notation}

The table below follows the problem definition and summarizes the notation for memory construction, retrieval, and complexity analysis.
MaC constructs $\mathcal{M}$ from dialogue history $\mathcal{H}$, executes a program $P_q$ generated for query $q$ to assemble evidence $E_q$, and generates an answer $a_q$ while recording an execution trace $\mathcal{T}_q$.
Local variables used only in Algorithm~\ref{alg:mac-framework} are introduced alongside the algorithm.

\begin{center}
\begin{minipage}{\linewidth}
\centering
\small
\textbf{Notation used throughout the paper}
\par\smallskip
\begin{tabular}{@{}p{0.29\linewidth}p{0.66\linewidth}@{}}
\toprule
Symbol & Meaning \\
\midrule
\multicolumn{2}{@{}l}{\textit{Task notation}} \\
$\mathcal{H}$, $s_j$, $n_{\mathrm{sess}}$ & Dialogue history, its $j$th session, and the number of sessions. \\
$\mathcal{M}$ & External memory constructed from $\mathcal{H}$. \\
$q$, $E_q$, $a_q$ & Query, supporting evidence assembled for it, and the generated answer. \\
\midrule
\multicolumn{2}{@{}l}{\textit{Codified memory}} \\
$\mathcal{L}$, $I$ & Code library and memory index, with $\mathcal{M}=(\mathcal{L},I)$. \\
$\mathcal{F}$, $\mathcal{V}$ & Base fact functions and views referencing base and derived functions. \\
$\mathcal{R}$, $\mathcal{D}$ & Explicit memory relations and derived memory functions. \\
$I_{\mathrm{view}}$, $I_{\mathrm{func}}$, $I_{\mathrm{ops}}$ & View-to-identifier mapping, function descriptors, and operator interfaces. \\
$i$, $f_i$ & Canonical fact identifier and its base fact function. \\
$e_i$, $v_i$, $S_i$ & Evidence record, structured value, and supporting source references for fact $i$. \\
$G$, $G(\tau)$, $A_G(\tau)$ & Linked state group, its temporally eligible fact identifiers, and its active fact identifiers. \\
$\tau$, $\bot$ & Requested time boundary and the marker for an unresolved state. \\
\midrule
\multicolumn{2}{@{}l}{\textit{Declarative retrieval}} \\
$K_q$ & Candidate function identifiers supplied to the planner for $q$. \\
$I_q(\mathcal{M})$ & Compact index context supplied to the planner for $q$. \\
$P_q$, $\mathcal{T}_q$ & Retrieval program for $q$ and its execution trace. \\
$F_{\mathrm{plan}}$, $F_{\mathrm{answer}}$ & Model stages for program generation and evidence-grounded answering. \\
\midrule
\multicolumn{2}{@{}l}{\textit{Complexity analysis}} \\
$n$, $h$ & Number of facts before deduplication and source-dialogue length in tokens. \\
$p$, $b$ & Syntax-tree size of $P_q$ and maximum expanded evidence size, counting scalar entries and text tokens. \\
$r$ & Stored references across views, relations, derived functions, and source annotations. \\
$T_{\mathrm{build}}$, $T_q$ & Construction time and total processing time for query $q$. \\
$\mathcal{C}_{\mathrm{build}}$, $\mathcal{C}_q$ & Model inference costs during construction and query processing. \\
\bottomrule
\end{tabular}
\end{minipage}
\end{center}

\subsection{Memory Operators}
\label{app:memory-operators}

Table~\ref{tab:mac2-operator-classification} summarizes the operators exposed to MaC search programs.

\begin{table*}[ht]
\centering
\caption{Memory operators exposed to MaC search programs.}
\label{tab:mac2-operator-classification}
\small
\renewcommand{\arraystretch}{1.12}
\setlength{\tabcolsep}{3pt}
\begin{tabularx}{\textwidth}{@{}
>{\centering\arraybackslash}m{0.16\textwidth}
>{\raggedright\arraybackslash\scriptsize}m{0.20\textwidth}      
>{\raggedright\arraybackslash}X@{}}    
\toprule    
Category & Core Operation & Description \\    
\midrule    
\multirow{6}{*}{\parbox[c]{0.16\textwidth}{\centering\textbf{Memory Access}}}      
& \texttt{resolve} & Executes selected memory functions. \\      
& \texttt{match} & Matches candidate facts to query keywords without code retrieval. \\      
& \texttt{search} & Ranks and retrieves canonical facts using a natural-language query. \\      
& \texttt{select} & Filters facts by structured constraints. \\      
& \texttt{search\_functions} & Finds memory functions matching the query constraints. \\      
& \texttt{execute} &  Runs a memory function and returns evidence. \\    
\midrule    
\multirow{3}{*}{\parbox[c]{0.16\textwidth}{\centering\textbf{Context Expansion}}}      
& \texttt{expand\_scope} & Adds related memories from the same concept or session. \\      
& \texttt{expand\_relation} & Follows graph links to add temporally or semantically related facts. \\      
& \texttt{hydrate} & Loads canonical facts and nearby dialogue for answer context. \\    
\midrule    
\multirow{4}{*}{\parbox[c]{0.16\textwidth}{\centering\textbf{Evidence Composition}}}      
& \texttt{union} & Merges evidence with its facts and dialogue sources. \\      
& \texttt{rank} & Orders evidence by relevance to the query. \\      
& \texttt{where} & Filters evidence by a specified time range. \\      
& \texttt{aggregate} & Counts evidence while retaining its facts and sources. \\    
\midrule    
\multirow{1}{*}{\parbox[c]{0.16\textwidth}{\centering\textbf{Verification}}}      
& \texttt{validate} & Checks evidence completeness and unresolved states. \\    
\bottomrule  
\end{tabularx}
\end{table*}

\subsection{Complexity Analysis}
\label{app:complexity}

This subsection expands Section~\ref{sec:complexity}.
To characterize the cost of reusable memory computation, we distinguish one-time construction from per-query retrieval and account for model inference separately.
Let $n$ be the number of extracted facts before deduplication, so $|\mathcal{F}|\leq n$, and let $h$ be the total length of the source dialogue $\mathcal{H}$ in tokens.
For a query $q$, let $p$ be the number of syntax-tree nodes in $P_q$ and $b$ the maximum fully expanded size of an evidence collection materialized during retrieval, counting scalar entries and text tokens in its values and provenance.
We assume bounded fact text, structured values, and query length, bounded annotations per extracted fact, and fixed retrieval and repair limits.

\paragraph{Time Complexity.}
Construction compares facts for deduplication and scans and sorts candidate neighbors for each canonical fact.
The latter costs $O(n^2\log n)$ in the worst case, even though only a bounded number of neighbors is retained per structural relation.
Including source preprocessing and model inference, the construction bound is $T_{\mathrm{build}}=O(h\log h+n^2\log n+\mathcal{C}_{\mathrm{build}})$.
During retrieval, candidate matching scans and sorts the memory index, while view expansion and provenance composition can incur quadratic work through list-based deduplication.
Structural program checking takes $O(p)$ time, and the loop-free language permits at most $O(p)$ operator calls.
Accounting for memory access, source loading, evidence composition, and trace generation gives the conservative bound $T_q=O\bigl(\mathcal{C}_q+p(n^2+h+b^2)+p^2\bigr)$.
Here $\mathcal{C}_{\mathrm{build}}$ and $\mathcal{C}_q$ sum the model inference costs in the corresponding stage, including extraction, semantic compilation, retrieval preprocessing, planning, and answer generation as applicable.
This accounting includes both operator execution and the model calls used across the query pipeline.

\paragraph{Space Complexity.}
Let $r$ count stored references across memory views, fact relations, derived functions, and source annotations.
The code library, memory index, construction records, and source dialogue require $O(n+r+h)$ space.
A query additionally retains intermediate evidence and its execution trace $\mathcal{T}_q$, giving $O(n+r+h+pb)$ space, excluding model weights and inference workspace.
Under the stated annotation bounds and the implemented relation limits, $r=O(n)$, so memory storage is linear in the input size.
The $pb$ term accounts for the generated program's intermediate results, including matching pairs materialized by joins and their provenance.

\subsection{Case Study (RQ4)}
\label{app:case-study}
\begin{figure}[htbp]
    \centering
    \includegraphics[width=\linewidth]{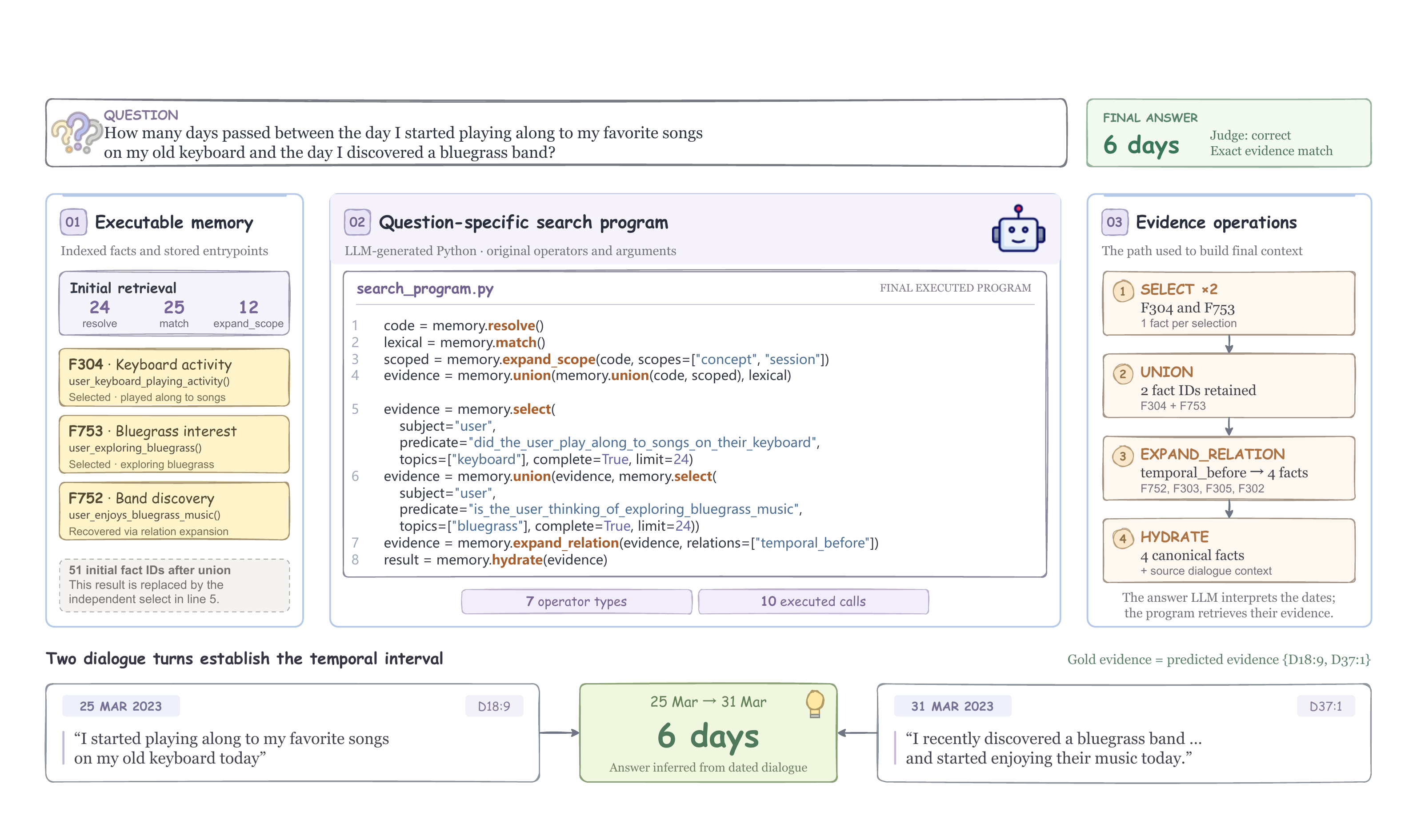}
    \caption{Temporal-reasoning case study on \textsc{LoCoMo}. MaC generates a question-specific retrieval program, expands temporal relations across sessions, and hydrates source dialogue; the answer model infers a \emph{6-day} interval matching the annotated evidence.}
    \label{fig:case-study}
\end{figure}

Figure~\ref{fig:case-study} presents a temporal-reasoning case in which the system must connect the user's keyboard-playing activity with a later bluegrass-band discovery across separate dialogue sessions.
MaC generates a question-specific search program that retrieves the relevant activity memories, expands their temporal relations to recover the event needed by the question, and restores the associated source dialogue for temporal grounding.
From the recovered evidence, the answer model correctly infers an interval of \emph{6 days}; its cited evidence exactly matches the annotated evidence.
This example demonstrates how programmatic memory search combines structured retrieval, relation-based expansion, and source-context recovery to answer temporally grounded conversational questions.

\section{Prompt details}
\label{app:prompt-details}

\begingroup
\begin{tcolorbox}[
    enhanced,
    breakable,
    sharp corners,
    boxrule=0.5pt,
    colback=white,
    colframe=black,
    label={box:fact_to_code},
    title=\textbf{Box 1: Fact to Code Prompt},
    fonttitle=\bfseries\sffamily,
    coltitle=black,
    attach boxed title to top left={yshift=-2mm, xshift=2mm},
    boxed title style={colback=white, colframe=white},
    drop shadow
]
\small

\textbf{Task:} Convert canonical facts into structured specifications that a local generator turns into executable Python memory functions.

\textbf{Input:} Canonical facts with identifiers, factual content, subjects, temporal information, and source references.

\textbf{System prompt:}

You are a semantic compiler. Produce one function specification for each fact while preserving its identifier and meaning. Specify a concise Python function name, a query predicate, the factual value, optional attributes, and whether the fact describes a time-varying state.

Use only the supplied facts. Do not invent information or infer state-update links in this stage. The specifications will be converted locally into Python functions.

\textbf{User prompt template:}

\textbf{Facts:}\\
\texttt{\char123\char123 facts\char125\char125}

Compile these facts into function specifications.

\textbf{Output:} A JSON object containing the fact specifications, not Python source code.

\end{tcolorbox}
\endgroup

\begingroup
\begin{tcolorbox}[
    enhanced,
    breakable,
    sharp corners,
    boxrule=0.5pt,
    colback=white,
    colframe=black,
    label={box:prompt_spec},
    title=\textbf{Box 2:  Declarative Query Planning Prompt},
    fonttitle=\bfseries\sffamily,
    coltitle=black,
    attach boxed title to top left={yshift=-2mm, xshift=2mm},
    boxed title style={colback=white, colframe=white},
    drop shadow
]
\small

\textbf{Task:} Generate a complete Python retrieval program that selects and composes memory operations for the given question.

\textbf{Input:} The question, compact candidate-function metadata, available memory interfaces, and general human-designed retrieval guidance.

\textbf{System prompt:}

Write one complete program over the read-only executable-memory SDK. Determine the required functions, operations, and dependencies from the question and interface information. Plan the retrieval as a whole rather than requesting further LLM decisions after each operation.

Use the human-designed guidance as reference, not as a fixed pipeline: direct lookup may suffice for a specific fact; distributed evidence may require expansion and composition; temporal questions may require state or history computations and source context. These patterns may not fit every question; adapt or omit them according to the actual evidence requirements.

Use compatible operations to retrieve, execute, filter, and combine evidence as needed. Preserve source information and assign the supporting evidence to \texttt{result}. Do not modify memory or generate the final natural-language answer.

\textbf{User prompt template:}

\textbf{Question:}\\
\texttt{\char123\char123 question\char125\char125}

\textbf{Candidate functions:}\\
\texttt{\char123\char123 functions\char125\char125}

Generate a complete retrieval program for this question using the available interfaces.

\textbf{Output:} A Python program whose \texttt{result} contains supporting memory evidence.

\end{tcolorbox}
\endgroup

\section{Framework Algorithm}
\label{app:framework-algorithm}

Algorithm~\ref{alg:mac-framework} uses the memory and query notation defined in Section~\ref{sec:method}.
The local collections $\mathcal{X}_{\mathrm{raw}}$ and $\mathcal{X}$ contain extracted and canonical fact records, respectively; a canonical record $x_i$ supplies the value $v_i$ and source references $S_i$ encoded by $f_i\in\mathcal{F}$.
For a query $q$, $K_q^{\mathrm{view}}$ contains function identifiers obtained through view expansion, and $E_q^{\mathrm{code}}$, $E_q^{\mathrm{view}}$, and $E_q^{\mathrm{lex}}$ denote evidence from code recall, view expansion, and lexical recall.
Their merged evidence is cached in $\mathcal{B}_q$; the program returns intermediate evidence $E$, which is assembled into the final evidence $E_q$.
Here $\mathcal{G}$ denotes the LLM used for extraction, annotation, and repair, while $F_{\mathrm{plan}}$ and $F_{\mathrm{answer}}$ denote the planning and answering stages defined in the main text.
Only values used by subsequent steps are shown.

\SetKwInput{KwIn}{Input}
\SetKwInput{KwOut}{Output}
\SetAlgoNlRelativeSize{-1}
\SetNlSty{textbf}{}{:}
\SetAlFnt{\small}
\SetAlCapSkip{0.35em}
\SetAlgoInsideSkip{0.25em}
\DontPrintSemicolon
\newcommand{\stage}[1]{\textcolor{green!50!black}{\textit{$\triangleright$ #1}}}

\begin{algorithm}[H]
\caption{Memory as Code (MaC) Framework}
\label{alg:mac-framework}
\KwIn{Dialogue history $\mathcal{H}=(s_1,\ldots,s_{n_{\mathrm{sess}}})$, query set $Q$, LLM $\mathcal{G}$}
\KwOut{Codified memory $\mathcal{M}$, answers $\{a_q\mid q\in Q\}$, and traces $\{\mathcal{T}_q\mid q\in Q\}$}
\stage{Stage 1: Codified Construction}\;
$\mathcal{X}_{\mathrm{raw}} \leftarrow \emptyset$\;
\ForEach{non-empty dialogue session $s_j\in\mathcal{H}$}{
  Extract source-annotated records with $\mathcal{G}.\mathrm{ExtractAtomicFacts}(s_j)$ and add them to $\mathcal{X}_{\mathrm{raw}}$\;
}
$\mathcal{X}\leftarrow\mathrm{CanonicaliseAndSemanticDeduplicate}(\mathcal{X}_{\mathrm{raw}})$; assign each record $x_i$ a canonical ID $i$\;
Store $\mathcal{X}$ in \texttt{facts\_by\_id.json}; initialize $\mathcal{R}$ with temporal, same-subject, and shared-topic relations\;
Annotate records in $\mathcal{X}$ using $\mathcal{G}$; group state facts by subject and predicate\;
Use $\mathcal{G}$ to identify explicit state replacements and add $(j,\texttt{supersedes},i)$ to $\mathcal{R}$\;
$\mathcal{F}\leftarrow\emptyset$\;
\ForEach{canonical record $x_i\in\mathcal{X}$}{
  Add to $\mathcal{F}$ a function $f_i$ returning $e_i=(v_i,\{i\},S_i)$; emit its applicable \texttt{supersedes} references\;
}
Construct views $\mathcal{V}$, derived current/history functions $\mathcal{D}$, and code for $\mathcal{R}$\;
Construct $\mathcal{L}=(\mathcal{F},\mathcal{V},\mathcal{R},\mathcal{D})$, $I=(I_{\mathrm{view}},I_{\mathrm{func}},I_{\mathrm{ops}})$, and $\mathcal{M}=(\mathcal{L},I)$\;
Persist \texttt{memory\_code}, \texttt{function\_index.json}, and \texttt{view\_index.json}\;
\stage{Stage 2: Declarative Retrieval}\;
\ForEach{query $q\in Q$}{
  $K_q\leftarrow\mathrm{CodeRecall}(q,I_{\mathrm{func}},24)$ \textbf{in parallel with}
  $\mathrm{terms}_q\leftarrow\mathcal{G}.\mathrm{ExtractRetrievalTerms}(q)$\;
  $E_q^{\mathrm{code}}\leftarrow\mathrm{ExecuteAndRehydrate}(K_q,\mathcal{M})$; $E_q^{\mathrm{lex}}\leftarrow\mathrm{KeywordRecall}(\mathrm{terms}_q,\texttt{facts\_by\_id},32)$\;
  $K_q^{\mathrm{view}}\leftarrow\mathrm{ViewExpandAndRank}(K_q,I_{\mathrm{view}},I_{\mathrm{func}},q,32)$;
  $E_q^{\mathrm{view}}\leftarrow\mathrm{ExecuteAndRehydrate}(K_q^{\mathrm{view}},\mathcal{M})$\;
  $\mathcal{B}_q\leftarrow\mathrm{Deduplicate}(E_q^{\mathrm{code}}\cup E_q^{\mathrm{view}}\cup E_q^{\mathrm{lex}})$\;
  Form $I_q(\mathcal{M})$ from $\{I_{\mathrm{func}}(\kappa)\mid\kappa\in K_q\}$ and $I_{\mathrm{ops}}$\;
  $P_q\leftarrow F_{\mathrm{plan}}(q,I_q(\mathcal{M}))$\;
  \For{$\ell\leftarrow0$ \KwTo $2$}{
    \If{$\mathrm{ValidatePlan}(P_q)$ and $\mathrm{RunInNetworkIsolatedSandbox}(P_q,\mathcal{M},\mathcal{B}_q)$ succeed}{
      $E\leftarrow\mathrm{ExecutionResult}(P_q)$; \textbf{break}\;
    }
    $P_q\leftarrow\mathcal{G}.\mathrm{RepairProgram}(P_q)$\;
  }
  \If{no generated program succeeds}{
    $P_q\leftarrow\mathrm{DefaultFallbackProgram}(I_q(\mathcal{M}))$; $E\leftarrow\mathrm{RunInNetworkIsolatedSandbox}(P_q,\mathcal{M},\mathcal{B}_q)$\;
  }
  $E_q\leftarrow\mathrm{RenderEvidenceContext}(E,\mathcal{B}_q,\mathcal{M})$; $a_q\leftarrow F_{\mathrm{answer}}(q,E_q)$\;
  \If{$a_q$ is malformed, empty, or a refusal}{ $a_q\leftarrow\mathcal{G}.\mathrm{RepairAnswer}(q,E_q,a_q)$\; }
  Record $P_q$, $E_q$, $a_q$, and line-level provenance in $\mathcal{T}_q$\;
}
\Return $\mathcal{M}$, $\{a_q\mid q\in Q\}$, and $\{\mathcal{T}_q\mid q\in Q\}$\;
\end{algorithm}

\end{document}